\documentclass{article}
\usepackage{amssymb}
\usepackage{amsmath}
\usepackage{graphicx}
\usepackage{enumitem}
\usepackage{hyperref}
\usepackage[preprint]{corl_2025} 
\usepackage{booktabs}
\usepackage{multirow}
\usepackage{colortbl}
\usepackage{rotating}
\usepackage[table]{xcolor}
\usepackage{wrapfig}
\usepackage{makecell}
\usepackage{threeparttable}
\hypersetup{
    colorlinks=true,
    linkcolor=orange,
    filecolor=magenta,
    urlcolor=orange,
    citecolor=orange,
}

\title{VTLA: Vision-Tactile-Language-Action Model with Preference Learning for Insertion Manipulation}

\author{
    Chaofan Zhang$^{1,}$\textsuperscript{*}, Peng Hao$^{2,}$\textsuperscript{*}, Xiaoge Cao$^{1}$, Xiaoshuai Hao$^{3}$, Shaowei Cui$^{1,}$\textsuperscript{†}, and Shuo Wang$^{1}$ \\
    \\
    \vspace{-2pt}
    $^{1}$State Key Laboratory of Multimodal Artificial Intelligence Systems, \\ 
    Institute of Automation, Chinese Academy of Sciences\\
    $^{2}$Samsung R\&D Institute China–Beijing \\
    $^{3}$Beijing Academy of Artificial Intelligence\\
}

\begin{document}
\maketitle

\vspace{-20pt}

\begin{figure*}[h] 
\setlength{\abovecaptionskip}{0cm}
    \centering
    \includegraphics[width=\textwidth]{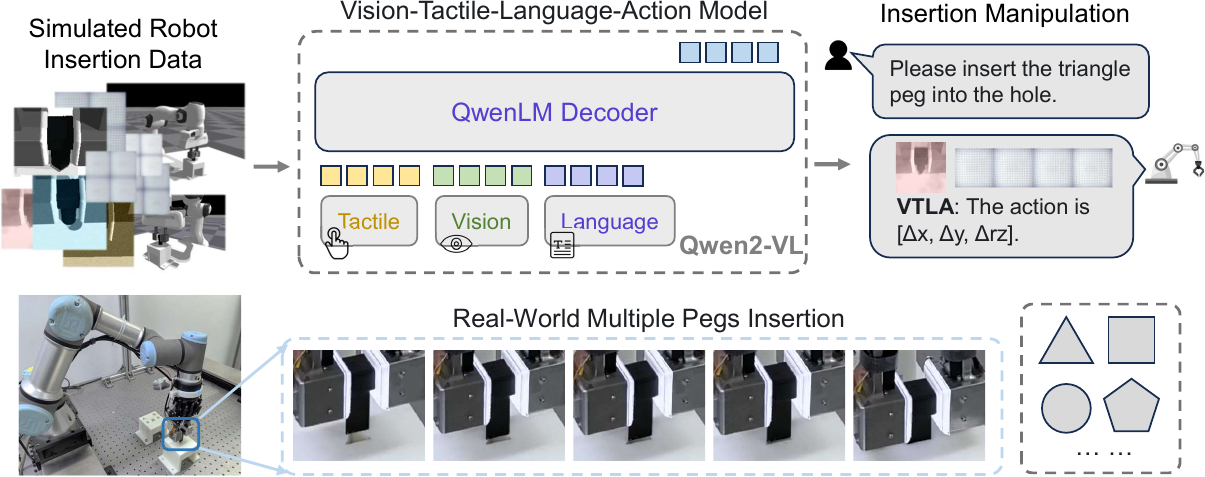}
    \caption{
    \textbf{Overview of VTLA.} The VTLA model learns a robotic manipulation policy integrated with vision, tactile, and language inputs from domain-randomized simulation data, enabling it to perform a variety of peg-in-hole tasks in the real world.
    }
\label{intro}
\end{figure*}
    
\vspace{-10pt}
\begin{abstract}
    While vision-language models have advanced significantly, their application in language-conditioned robotic manipulation is still underexplored, especially for contact-rich tasks that extend beyond visually dominant pick-and-place scenarios.
    To bridge this gap, we introduce \underline{\textbf{V}}ision-\underline{\textbf{T}}actile-\underline{\textbf{L}}anguage-\underline{\textbf{A}}ction (\textbf{\textit{VTLA}}) model, a novel framework that enables robust policy generation in contact-intensive scenarios by effectively integrating visual and tactile inputs through cross-modal language grounding.
    A low-cost, multi-modal dataset has been constructed in a simulation environment, containing vision-tactile-action-instruction pairs specifically designed for the fingertip insertion task.
    Furthermore, we introduce Direct Preference Optimization (\textit{DPO}) to offer regression-like supervision for the \textbf{\textit{VTLA}} model, effectively bridging the gap between classification-based next token prediction loss and continuous robotic tasks.
    Experimental results show that the \textbf{\textit{VTLA}} model outperforms traditional imitation learning methods (\textit{e.g.}, diffusion policies) and existing multi-modal baselines (TLA/VLA), achieving over 90\% success rates on unseen peg shapes.
    Finally, we conduct real-world peg-in-hole experiments to demonstrate the exceptional Sim2Real performance of the proposed \textbf{\textit{VTLA}} model.
    For supplementary videos and results, please visit our project website: \href{https://sites.google.com/view/vtla}{VTLA}.
    
\end{abstract}
\vspace{-10pt}
\keywords{Contact-Rich Manipulation, Tactile Sensing, Large Language Model} 

\renewcommand{\thefootnote}{} %
\footnote{%
\hangindent=1.8em\textsuperscript{*}Equal contribution\\
\textsuperscript{†}Corresponding author: shaowei.cui@ia.ac.cn%
}

\vspace{-30pt}
\section{Introduction}
\vspace{-10pt}
In contact-intensive manipulation tasks, such as precise object insertion~\cite{hansen2022visuotactile,lee2020making}, vision is essential for environmental perception. 
However, humans naturally integrate tactile feedback to manage uncertainties~\cite{hogan2020tactile,wu2024tacdiffusion,castano2024visual}. 
For example, aligning a key to a lockhole with an ambiguous position necessitates reliance on tactile feedback to compensate for limited visual information~\cite{wang2025temptrans,george2024vital,billard2019trends}.
This highlights the importance of vision-tactile fusion in achieving robust robotic manipulation~\cite{cui2021toward}.
The recently emerging vision-tactile learning frameworks offer a promising foundation for robotic skill acquisition, allowing robots to interpret complementary contact-state signals to enhance grasping and manipulation~\cite{calandra2018more,cui2020self,feng2024play}.
However, existing studies primarily rely on training proprietary models on specific datasets, leading to challenges in generalization across diverse scenarios and a lack of adaptability for human-like extensive perceptual-motor reasoning~\cite{xiao2025robot,xu2024survey,brohan2023rt}.

Recently, Large Language Models (LLMs) have made significant advancements in human-like reasoning~\cite{chang2024survey}. This progress has accelerated the development of Vision-Language-Action (VLA) models specifically designed for robotic manipulation~\cite{kim2024openvla, wen2025tinyvla, tang2025affordgrasp,li2024foundation}.
By aligning visual modalities with language modalities and leveraging LLMs for reasoning~\cite{ding2025humanoid, ji2025robobrain,tan2025reason, zhen20243d}, VLA models have significantly surpassed traditional imitation learning methods~\cite{ma2024hierarchical,li2024learning} in their ability to generalize across diverse robotic platforms and task configurations~\cite{gbagbe2024bi, zhu2025objectvla}.
However, the lack of tactile feedback in most VLA systems limits their functionality to simple tasks, such as grasping and placing. This restriction hinders their applicability in contact-rich manipulation scenarios~\cite{han2025multimodal}.

Tactile LLMs have demonstrated impressive understanding and reasoning capabilities in tactile perception, such as texture description and material recognition~\cite{tu2025texttoucher, cheng2024touch100k, feng2025anytouch, yang2024binding, yu2024octopi}. However, their application in language-conditioned action modeling remains nascent~\cite{jones2025beyond}.
A Tactile-Language-Action (TLA) model~\cite{hao2025tla} has been proposed for contact-rich manipulation tasks, demonstrating its potential in generalist tactile policy learning.
Further investigation reveals two key limitations: 1) the tactile encoding and training/inference framework of the TLA model still have significant room for improvement; and 2) the absence of visual modalities imposes performance ceilings, as the robot lacks global perception capabilities.

To address these limitations, we introduce the \underline{\textbf{V}}ision-\underline{\textbf{T}}actile-\underline{\textbf{L}}anguage-\underline{\textbf{A}}ction (\textbf{\textit{VTLA}}) framework, designed for effective contact-rich manipulation by integrating visual, tactile, and linguistic information.
The technical contributions of VTLA are twofold. First, we design Vision-Guided Temporally Enhanced Tokens (VGTE) based on the VLM capabilities and the characteristics of vision-tactile manipulation tasks. By emphasizing visual tokens and enhancing temporal fusion before tokenization, VGTE mitigates the limitations of VLMs in temporal understanding and improves VTLA's performance.
Second, we employ Direct Preference Optimization (DPO)~\cite{rafailov2023direct} to offer regression-like supervision for our action-conditional model.
To further evaluate the Sim2Real transferability of \textbf{\textit{VTLA}}, we created a vision-tactile robotic peg-in-hole assembly environment that replicates the simulation setup at a 1:1 scale. We then conducted real-world experiments on the VTLA model, which was trained exclusively on simulation data, across various peg-hole clearances and geometric configurations.

Our main contributions are summarized as follows:
\begin{itemize}[left=1.2em]
    \item We propose \textbf{\textit{VTLA}}, a novel vision-tactile fusion framework that integrates perception and language-action generation for contact-rich manipulation tasks.
    \item VGTE is designed to address the limited temporal reasoning capability of VLMs in vision-tactile manipulation. By emphasizing visual priors and incorporates temporal fusion prior to tokenization, VGTE enhances the cross-modal temporal reasoning of VTLA model.
    \item Preference Learning is introduced into VTLA to mitigate overfitting to ground-truth actions. By leveraging DPO to simulate regression-like supervision, VTLA gains richer training signals and performs enhanced generalization.
    \item Real-world insertion experiments show that \textbf{\textit{VTLA}} outperforms current methods, demonstrating the effectiveness of the designed modules. 
\end{itemize}

We hope this work will provide new insights for tactile-embedded VLA frameworks and inspire future research.

\section{Related Works}
\label{sec:related_works}

\textbf{Vision-Tactile Learning}
Vision-tactile fusion perception is crucial for improving robotic grasping and manipulation~\cite{liu2016visual}. With the advent of deep learning, various fusion mechanisms—such as early feature fusion~\cite{cui2020grasp} and attention mechanisms~\cite{cui2020self}—have been explored for tasks like slip detection~\cite{li2018slip}, grasp outcome prediction~\cite{calandra2018more}, and liquid pouring~\cite{feng2024play}. Reinforcement learning methods leveraging vision-tactile fusion have been proposed for peg-in-hole assembly~\cite{hansen2022visuotactile,lee2020making}. Recently, the combination of vision-tactile data acquisition with imitation learning has emerged as a promising area~\cite{liu2025vitamin,xue2025reactive,zhang2025mapnav,huang20243d,wuevaluating,hong2024multiply}. However, these approaches often depend on specialized models trained on fixed datasets, limiting their out-of-domain generalization compared to general-purpose models.

\textbf{VLM for Robot Manipulation}
Recent advancements in robotic manipulation showcase the potential of vision-language models (VLMs) with enhanced reasoning capabilities. RT-2~\cite{brohan2023rt} formulates robotic actions as token sequences through fine-tuning on manipulation datasets, establishing the Vision-Language-Action (VLA) paradigm. 
Subsequent studies, such as RoboFlamingo~\cite{li2023vision} and OpenVLA~\cite{kim2024openvla}, adopt similar methods.
GR-1~\cite{wu2023unleashing} and GR-2~\cite{cheang2024gr} employ a two-stage training approach, first pre-training on web-scale video corpora and then adapting to manipulation datasets. Innovations like RDT-1B~\cite{liu2024rdt} and PI~\cite{black2024pi_0} integrate stochastic modeling with diffusion objectives to enhance action sequence generation. However, existing VLA models mainly focus on visual modalities, limiting their use to relatively simple tasks like planar pushing and pick-and-place operations. This study investigates the integration of tactile perception into a vision-language foundation model, targeting contact-rich manipulation challenges such as peg-in-hole insertion.

\textbf{Tactile-Language Model in Robotics}
Recent works in tactile-language model emphasize material understanding. Fu et al.~\cite{fu2024touch} created a ChatGPT-assisted tactile-texture dataset using Tactile-Vision-Language Models, while Cheng et al.~\cite{cheng2024touch100k} expanded this and presented Touch100k. Despite these advancements, tactile data is still underutilized in robotic manipulation. Most recently, Jones et al.~\cite{jones2025beyond} embed tactile signals into VLA models for interaction policies. TLA~\cite{hao2025tla} links tactile modalities and language to generate robot actions, indicating the potential of tactile modalities in language-conditioned robot learning. 
In this paper, we propose VTLA model to generate robotic manipulation policies by integrating visual, tactile, and language modalities.
	

\section{Vision-Tactile-Language-Action Model}
\label{sec:methods}

\subsection{Data Collection}
\begin{figure*}[t] 
\setlength{\abovecaptionskip}{0cm}
    \centering
    \includegraphics[width=\textwidth]{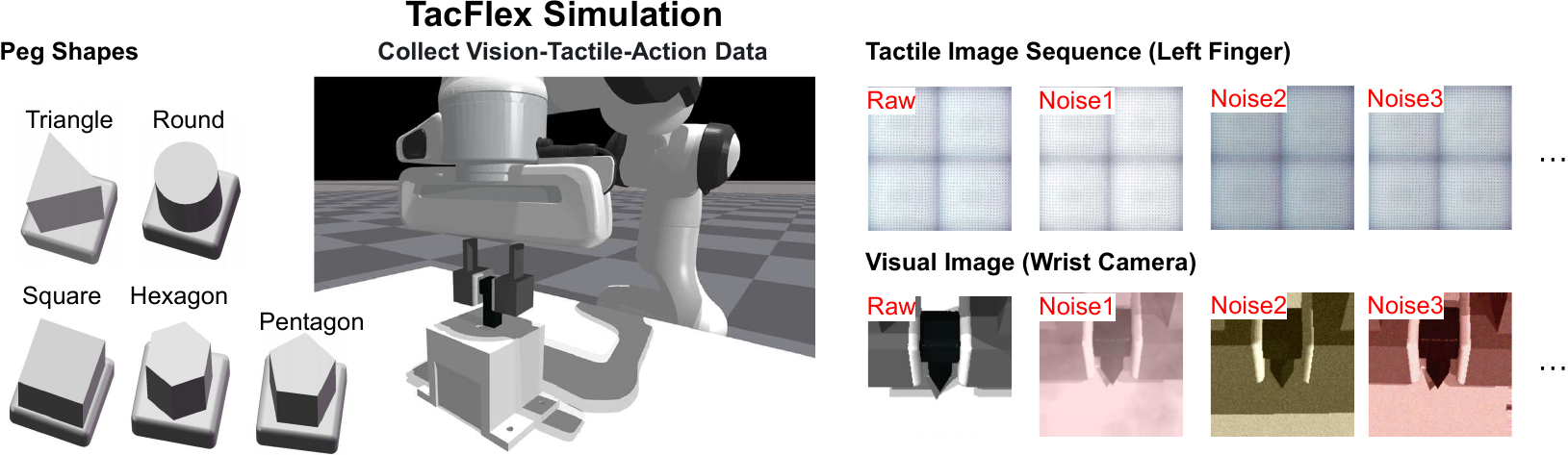}
    \caption{The data collection diagram and data examples of the VTLA dataset. }
\label{dataset}
\vspace{-12pt}
\end{figure*}
To enhance data collection efficiency and evaluate the Sim2Real generalization capability of the model, we adopt a synthetic data training approach followed by real-world validation. We construct a peg-in-hole assembly task in NVIDIA Isaac Gym with a self-built visuotactile simulator.
This setup integrates a wrist-mounted camera and visuotactile sensors on the gripper fingertips to capture both visual and tactile observations during the assembly process, as illustrated in Fig.~\ref{dataset}.

The assembly task is structured as follows: The gripper grasps a peg, positions it above the corresponding hole, and introduces a randomized 3-DOF misalignment in the x-axis, y-axis, and z-axis rotation. The gripper then descends to attempt insertion. If a collision occurs, the attempt fails, and the gripper retracts for another try. If no collision happens before reaching the insertion depth, the task is successful. The maximum number of attempts is set to 15; otherwise, the task fails. In simulation, a randomized insertion strategy generates visual and tactile data, with task configurations and action labels aligned with those in TLA~\cite{hao2025tla}.

We collect data for five distinct peg-hole shapes with assembly clearances from 0.6 to 2.0 mm. The VTLA dataset comprises 28,000 assembly samples, each containing left/right tactile image sequences, a visual image, and an action label. The tactile image sequences are arranged in a $2 \times 2$ grid, as shown in Fig. \ref{dataset}. To enhance zero-shot Sim2Real transfer performance, domain randomization techniques are employed during dataset generation to vary parameters in the simulation environment, task configurations, and both visual and tactile observations. Details on domain randomization are provided in the supplementary materials.

To facilitate training of the \textit{\textbf{VTLA}} model, the dataset is organized in an instruction format. The \texttt{<|im\_start|>} and \texttt{<|im\_end|>} tokens mark the start and end of each dialogue round. Tactile and visual images are input sequentially with \texttt{<|vision\_start|>} and \texttt{<|vision\_end|>}. A text instruction specifies the task, including image types, peg shapes, and robot action requirements, while the action label serves as the ground truth. An example of \textit{\textbf{VTLA}} data is provided below.

\begin{figure*}[h] 
\setlength{\abovecaptionskip}{0cm}
    \centering
    \includegraphics[width=\textwidth]{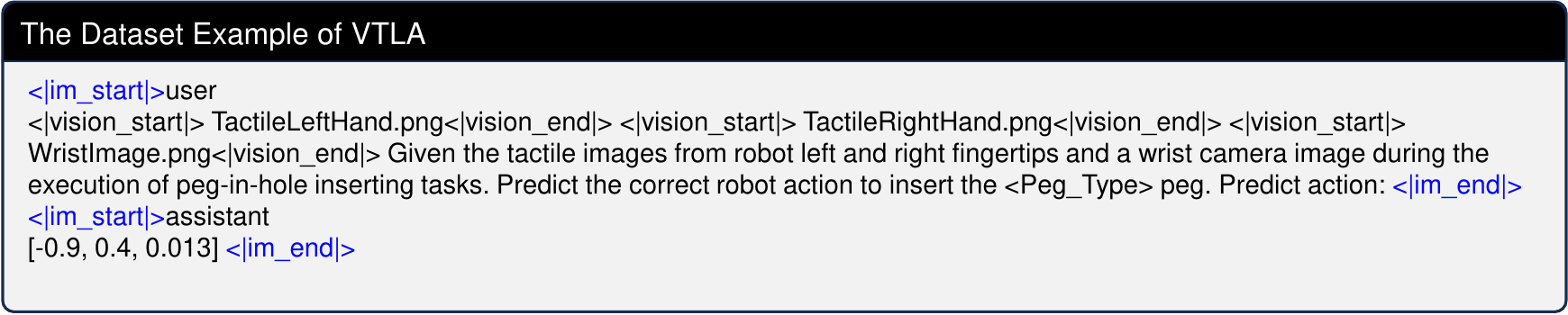}
\vspace{-18pt}
\end{figure*}

\subsection{Instruction Tuning with Vision-Guided Temporally Enhanced Tokens}

\textbf{Vision-Guided Temporally Enhanced Tokens} While VLMs demonstrate superior generalization, their performance on specific tasks remains sensitive to the input prompts. This section presents the design of VTLA, which is tailored to the characteristics of both language models and visual-tactile manipulation tasks. VTLA introduces a vision-guided temporal enhancement mechanism to construct multimodal tokens that serve as temporally aligned inputs for the language model. This design enables more effective cross-modal reasoning by leveraging the multimodal understanding of pre-trained VLM, thereby facilitating superior performance in visual-tactile control task.

VTLA’s multimodal inputs focus on two key aspects: vision guidance and temporal enhancement. Prior studies \cite{lee2020making} show that visual observations are crucial in the early stages of visual-tactile tasks. To reflect this priority and address the recency bias of language models \cite{peysakhovich2023attention}, VTLA positions visual inputs after tactile inputs, bringing vision closer to action prediction. This strategy emphasizes the importance of visual information during initial manipulation phases, enabling VTLA to better utilize essential visual cues for contact-rich control.

Furthermore, VTLA employs temporally enhanced tactile inputs to improve the VLM’s reasoning of sequential data. While VLMs excel in cross-modal reasoning, they struggle with fine-grained temporal dependencies \cite{fei2024enhancing}. Although the Qwen-VL family uses 3D convolutions for video processing \cite{qwen2-vl,bai2025qwen2}, there is a domain gap between their high-level semantic tasks and the short-duration, low-level nature of tactile images in robotic manipulation. To bridge this gap, we encode tactile observations into image-like representations and extract temporally-aware tactile tokens using a Vision Transformer (ViT). This approach addresses the temporal reasoning limitations of VLMs and leverages their strengths in multimodal comprehension.

\begin{figure*}[t] 
\setlength{\abovecaptionskip}{0cm}
    \centering
    \includegraphics[width=0.85\textwidth]{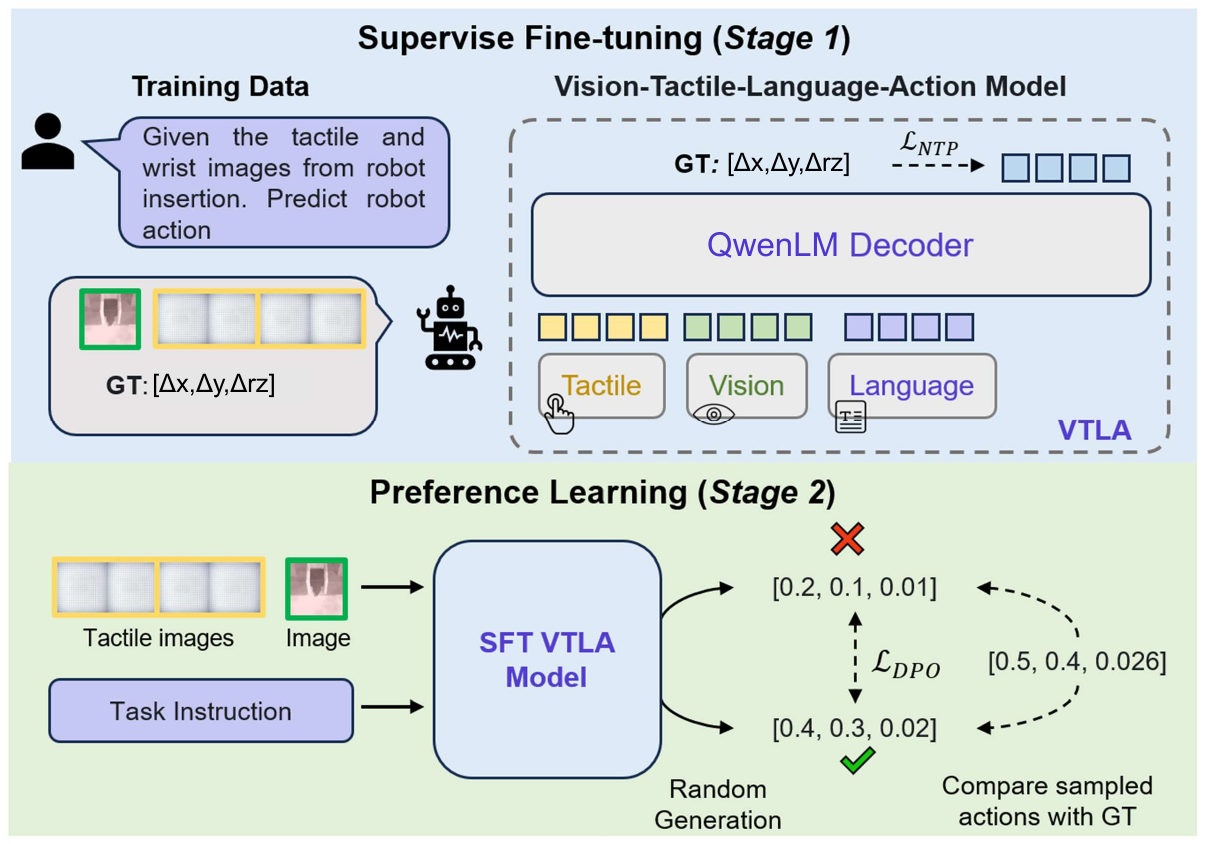}
    \caption{\textbf{The pipeline of VTLA.} 
    In stage 1, the instruction dataset is created from simulation data using a vision-guided temporal enhancement, and the VTLA model is optimized with NTP loss. In stage 2, DPO is introduced to provide regression-like supervision, bridging the gap between VLM training and robotic continuous control, thereby enhancing performance.}
\label{fig-pipeline}
\vspace{-10pt}
\end{figure*}

\textbf{Supervised Fine-Tuning} Guided by the above two designs, we fine-tune the VTLA model with the collected simulation dataset. Following prior works~\cite{kim2024openvla,hao2025tla}, we formulate VTLA as a Next Token Prediction (NTP) task as shown in Fig.~\ref{fig-pipeline}. Specifically, tactile and visual observations are first processed through a pre-trained vision encoder and a modality adapter to generate tactile and vision tokens. Meanwhile, the textual instruction is tokenized to obtain text tokens. These generated tactile, vision, and text tokens then fed into a pre-trained LLM to predict the robot action. The entire model is fine-tuned using the NTP loss, defined as follows,
\begin{equation}
    \label{eq-ntploss}
    \mathcal{L}_{\text{NTP}} = - \sum_{n=1}^{N} \log P_\theta(x_n \mid x_{<n}),
\end{equation}
\noindent where $N$ denotes the total length of input tokens, $x_n$ is the ground-truth token at position $n$, and $x_{<n}$ represents the preceding tokens as context. $\theta$ denotes the trainable parameters of the VTLA model. Following prior works~\cite{karamcheti2024prismatic}, we freeze the parameters of the vision encoder and modality adapter, tuning only the language model to enhance performance.

\subsection{Preference Learning}
The instruction-tuned VTLA model shows limited performance due to a mismatch between robotic control tasks and the classification-based NTP loss. Robotic control is a regression problem that requires predicting continuous control signals, but Stage 1 of the VTLA model uses a classification-oriented NTP loss, overlooking this aspect. To address this, we reformulate the VTLA prediction task as a multi-label problem, enabling richer supervision through multi-label optimization. We specifically introduce Direct Preference Optimization~\cite{rafailov2023direct} into VTLA, simulating a regression-like loss through preference learning.

The preference learning process is illustrated in Fig.~\ref{fig-pipeline}. We first use the fine-tuned VTLA model to generate diverse action predictions from the same training samples. Based on their proximity to the ground truth, we create a preference dataset, labeling actions closer to the ground truth as chosen and others as rejected. Finally, we optimize the fine-tuned model on this dataset using DPO. The training objective is as follows:
\begin{equation}
\mathcal{L}_{\text{DPO}} = - \log \sigma\left( \beta  \log \dfrac{\pi_\theta(y_{\text{chosen}} \mid x)}{\pi_{ref}(y_{\text{chosen}} \mid x)}  - \log \dfrac{\pi_\theta(y_{\text{rejected}} \mid x)}{\pi_{ref}(y_{\text{rejected}} \mid x)}\right),
\end{equation}
\noindent where $x$, $y_{\text{chosen}}$, and $y_{\text{reject}}$ represent the input tokens, preferred responses, and rejected responses, respectively. $\pi_{\theta}$ and $\pi_{\text{ref}}$ are the trainable and frozen VTLA models, both initialized from the fine-tuned model in stage 1. The scalar $\beta > 0$ controls the preference signal's sharpness, and $\sigma(\cdot)$ is the sigmoid function. This loss function promotes a higher likelihood for the preferred response over the rejected one, aligning the output with preference-based supervision.
	

\section{Experiments}
\label{sec:result}
We conduct a series of experiments to answer the following key research questions:
\textbf{(Q1)} How does our proposed method compare with existing state-of-the-art approaches?
\textbf{(Q2)} What is the impact of preference learning on model performance? 
\textbf{(Q3)} Can the proposed method generalize to real-world robotic applications?

\subsection{Experiment Setup}
\textbf{Existing and Ablation Methods}  
To address the above questions, we compare several baseline and ablation methods with the proposed VTLA on the dataset and various insertion tasks in simulation. 1) The \textbf{\textit{Diffusion Policy (DP)}} \cite{chi2023diffusion} learns the insertion policy using vision and tactile observations. 2) A \textbf{\textit{Vision-Language-Action (VLA)}} model is trained on wrist camera images from the dataset. 3) A \textbf{\textit{Tactile-Language-Action (TLA)}} model is trained on tactile image sequences. 4) We tested the fine-tuned VTLA model under two generation configurations, obtaining 2,400 preference data points by comparing with the ground truth. Models trained with 1,000 and 2,400 preference data are denoted as \textbf{\textit{VTLA~(w/ DPO-1k)}} and \textbf{\textit{VTLA~(w/ DPO-2k)}}, respectively.

\textbf{Evaluation Metrics} 
For dataset evaluation, we define the \texttt{Goal Convergence Rate (GCR)} as the percentage of actions that are all correct in the $x$, $y$, and $rz$ directions. The \texttt{L1 distance} between output actions and action labels is used to evaluate performance in each direction. For the insertion task in simulation, we compute the success rate and average attempt steps for each method.

\textbf{Real-World Insertion Robot Setup} 
We use a 6-DoF UR3 robot arm with a Robotiq 2F-85 gripper for real-world insertion experiments. An Intel RealSense D405 camera is mounted on the wrist to capture vision images, while two GelStereo 2.0 sensors \cite{zhang2023gelstereo} on the gripper's fingertips are used to obtain tactile observations during peg-hole collisions.

\textbf{Implementation Details} 
We utilized the \texttt{LlamaFactory} framework~\cite{zheng2024llamafactory} for both SFT and DPO training. During the SFT stage, we used the \texttt{Qwen2-VL 7B} model as the base model, with a learning rate of \(5 \times 10^{-4}\), a batch size of 64, and 10 epochs. In the DPO stage, the SFT model are served as the initialized model, with a learning rate of \(5 \times 10^{-6}\), a batch size of 32, and 3 epochs.

\subsection{Comparison with Baseline Methods} 

\begin{table}[h]
   \vspace{-0.5em}
    \setlength{\abovecaptionskip}{-0.2cm}
    \caption{Comparison of different methods on the dataset.}
    \label{tab_dataset_1}
    \centering
    \resizebox{0.69\textwidth}{!}{ 
    \setlength{\tabcolsep}{0.7mm}{
    \begin{tabular}{l|cccc|cccc}
    \hline\hline
    \multirow{2}{*}{Method}                 & \multicolumn{4}{c|}{ID}                                & \multicolumn{4}{c}{OOD}                                \\
               & GCR(\%) &  L1 x & L1 y & L1 rz & GCR(\%)  &  L1 x & L1 y & L1 rz \\ \hline
    DP~\cite{chi2023diffusion}               &  7.8          &   0.826       &  0.819    &   1.421    &  8.5          &   0.821       & 0.843     & 1.407      \\
    VLA~\cite{qwen2-vl}             &  \underline{46.1}           &   \underline{0.210}       &     \underline{0.247} &   \textbf{0.886}    &  \underline{29.5}          & \underline{0.353}         &  \underline{0.351}    & \underline{1.221}      \\
    TLA~\cite{hao2025tla}              &  15.3           &  0.531        &  0.677    & 1.427      & 14.4          &   0.509       &  0.675    &   1.462    \\
   \rowcolor[HTML]{DAEFF9} \textbf{VTLA (Ours)}     & \textbf{47.3}  & \textbf{0.181} & \textbf{0.224} & \underline{0.904} & \textbf{31.2} & \textbf{0.305} & \textbf{0.324} & \textbf{1.136} \\
    \hline\hline
    \end{tabular} 
    }}
    \vspace{-10pt}
\end{table}

\begin{wraptable}{r}{7.5cm}
    \setlength{\abovecaptionskip}{-0.4cm}
    \caption{Comparison of different methods on the \textbf{square peg} insertion tasks with different clearances in simulation.}
    \label{tab_sim_exp1}
    \centering
    \resizebox{0.54\textwidth}{!}{
    \setlength{\tabcolsep}{0.7mm}{
    \begin{tabular}{l|cccccccc}
    \hline\hline
    \multirow{2}{*}{Method}                 & \multicolumn{2}{c}{2.0 mm} & \multicolumn{2}{c}{1.6 mm} & \multicolumn{2}{c}{1.0 mm} & \multicolumn{2}{c}{0.6 mm} \\
                     & Suc         & Step         & Suc         & Step         & Suc         & Step         & Suc         & Step         \\ \hline
    DP~\cite{chi2023diffusion}               &    42         &  2.47            &  32           &  2.63            &        28     &   4.85           &    22         &    3.54       \\
    VLA~\cite{qwen2-vl}              &    \textbf{100}        &   2.28           &    \textbf{98}      &    3.24        &    90       &    3.28      &                 80        &   5.55         \\
    TLA~\cite{hao2025tla}              &     94       &   3.27           &   90       &   3.60          &     80      &     4.97             &         80    &     5.48          \\
    \rowcolor[HTML]{DAEFF9} \textbf{VTLA (Ours)}              &  \textbf{100}          &    2.12          &     \textbf{98}     &   2.87          &    \textbf{96}       &    4.64              &       \textbf{90}      &    5.91      \\ \hline\hline
    \end{tabular}
    }}

\vspace{-10pt}
\end{wraptable}
We compare the performance of VTLA with baseline methods on the dataset. We take 6k samples from the In-Distribution (ID) set, and 4k samples from the Out-Of-Distribution (OOD) set for evaluation, and the quantitative results are shown in Tab.~\ref{tab_dataset_1}. Furthermore, we evaluate these methods on various insertion tasks within the simulation environment. Specifically, square pegs with clearance of 2.0, 1.6, 1.0, and 0.6 mm are tested, following with pegs of various shapes with 0.6 mm clearance. Each insertion task is tested 50 times in simulation.
The experimental results are presented in Tab.~\ref{tab_sim_exp1} and Tab.~\ref{tab_sim_exp2}.

The experimental results in Tab.~\ref{tab_dataset_1} and Tab.~\ref{tab_sim_exp1} show that the LLM-based models perform significantly better than the DP method in this insertion task (Tab.~\ref{tab_sim_exp1}: 80+\% vs 22\% success rate in the 0.6 mm clearance insertion experiment). 
The VTLA and VLA model significantly outperform the TLA model in quantitative results (Tab.~\ref{tab_dataset_1}:  GCR-ID, 46+\% vs 15.3\%). These findings highlight the key role of visual input in robot manipulation tasks. In addition, we find that VTLA with tactile encoding achieves better dataset performance (Tab.~\ref{tab_dataset_1}:  GCR-ID, 47.3\% vs 46.1\%) and task success rate (Tab.~\ref{tab_sim_exp1}:  0.6 mm clearance, 90\% vs 80\%) than VLA, which shows the effectiveness of temporally-aware tactile tokens in the proposed VTLA model.

\begin{wraptable}{r}{8.9cm}
    \setlength{\abovecaptionskip}{-0.5cm}
    \caption{Comparison of different methods on the peg-in-hole task\textbf{ with 0.6 mm clearance} in simulation.}
    \label{tab_sim_exp2}
    \centering
    \resizebox{0.65\textwidth}{!}{ 
    \setlength{\tabcolsep}{0.7mm}{
    \begin{tabular}{l|cccccc|cccc}
    \hline\hline
    \multirow{3}{*}{Method}                 & \multicolumn{6}{c|}{ID}                                                                  & \multicolumn{4}{c}{OOD}                                  \\
                    & \multicolumn{2}{c}{Square} & \multicolumn{2}{c}{Triangle} & \multicolumn{2}{c|}{Hexagon} & \multicolumn{2}{c}{Pentagon} & \multicolumn{2}{c}{Round} \\
                    & Suc         & Step         & Suc          & Step          & Suc          & Step          & Suc          & Step          & Suc         & Step        \\ \hline
    DP~\cite{chi2023diffusion}              &      22       &  3.54            &    30          &   3.87            &      28        &     3.00          &    26          &  5.61             &   10          & 3.80            \\
    VLA~\cite{qwen2-vl}             &   80        &  5.55            &      82        &   5.02           &        84      &    3.83         &   82          &     4.41          &   \textbf{94}          &     4.81        \\
    TLA~\cite{hao2025tla}             &   80        &   5.48           &    74          &   4.27           &          80    &  5.25             &   80           &    4.60           &       92      &    3.54         \\
    \rowcolor[HTML]{DAEFF9} \textbf{VTLA (Ours)}             &  \textbf{90}         &  5.91            &       \textbf{88}       &   4.53           &  \textbf{90}            &    4.68           &        \textbf{92}      &     3.97          & 92            &     4.74        \\
    \hline\hline
    \end{tabular}
    }}
\vspace{-10pt}
\end{wraptable}

In terms of generalization performance, the LLM-based models perform significantly better than the DP method (Tab.~\ref{tab_sim_exp2}: OOD-Round, 90+\% vs 10\%).
Furthermore, the VTLA model is better than TLA and VLA model in all dimensions of OOD data (Tab.~\ref{tab_dataset_1}: OOD-GCR, 31.2 vs 29.5/14.4). In the pentagonal peg insertion task, VTLA achieves a higher insertion success rate (92\% vs 82\%/80\%) and fewer steps (3.97 vs 4.41/4.60). 
Interestingly, all three models demonstrate comparable performance in the round peg insertion experiment, which may be attributed to the geometric isotropy of the round shape, making this task inherently less challenging.

\subsection{Ablation Study}

\begin{table}[h]
    \setlength{\abovecaptionskip}{-0.3cm}
    \caption{Ablation study on direct preference optimization.}
    \label{tab_e2_2}
    \centering
    \resizebox{0.76\textwidth}{!}{ 
    \setlength{\tabcolsep}{0.7mm}{
    \begin{tabular}{l|cccc|cccc}
    \hline\hline
    \multirow{2}{*}{Method}                 & \multicolumn{4}{c|}{ID}                                & \multicolumn{4}{c}{OOD}                                \\
               & GCR(\%)  &  L1 x & L1 y & L1 rz & GCR(\%) & L1 x & L1 y & L1 rz \\ \hline
    VTLA~(w/o DPO)  & \textbf{47.5}  & 0.184 & 0.227 & 0.907 & 27.0  & 0.349 & 0.367 & 1.223 \\
    VTLA~(w/ DPO-1k)   & \textbf{47.5}  & \textbf{0.181} & \textbf{0.224} & 0.906 & \textbf{31.4}  & \textbf{0.305} & \textbf{0.324} & 1.137 \\
    \rowcolor[HTML]{DAEFF9} VTLA~(w/ DPO-2k)     & 47.3  & \textbf{0.181} & \textbf{0.224} & \textbf{0.904} & 31.2  & \textbf{0.305} & \textbf{0.324} & \textbf{1.136} \\
    \hline\hline
    \end{tabular} 
    }}
    \vspace{-10pt}
\end{table}

The ablation study on DPO is presented in Tab.~\ref{tab_e2_2}. Experimental results show that preference learning with DPO significantly improves performance on both ID and OOD data. Specifically, VTLA-DPO-1k achieves a 16\% improvement in GCR and approximately a 10\% reduction in L1 error across all dimensions on OOD data. These improvements are attributed to the alignment of the DPO optimization objective with the nature of continuous robotic control tasks. By alleviating overfitting to sampled ground-truth actions during the SFT stage, DPO enhances the model’s generalization capabilities, particularly on OOD samples. Furthermore, we observe that increasing the size of the preference dataset does not lead to further performance gains. As the current preference data is generated by comparing outputs from only two sets of sampling configurations, we hypothesize that increasing diversity in preference data may be more effective than merely scaling dataset size.

\subsection{Real-world Robotic Insertion}

\begin{table*}[h]
    \centering
    \setlength{\tabcolsep}{0.5mm}

    \begin{minipage}{0.38\textwidth}
        \setlength{\abovecaptionskip}{0cm}
        \caption{Real-world insertion tasks on \textbf{square pegs} with different clearances.}
        \label{tab_e3_1}
        \resizebox{\textwidth}{!}{ 
        \begin{tabular}{c|cccccccc}
            \hline\hline
            &  \multicolumn{2}{c}{1.6 mm} & \multicolumn{2}{c}{1.0 mm} & \multicolumn{2}{c}{0.6 mm}\\
            & Suc         & Step         & Suc         & Step         & Suc         & Step         \\ \hline
            \rowcolor[HTML]{DAEFF9} VTLA   & 100         & 1.60       & 100         & 1.95           & 95         & 4.31       \\ \hline\hline
        \end{tabular}
        }
    \end{minipage}%
    \hfill
    \begin{minipage}{0.58\textwidth}
        \setlength{\abovecaptionskip}{0cm}
        \caption{Real-world insertion tasks on different pegs with \textbf{0.6 mm clearances}.}
        \label{tab_e3_2}
        \resizebox{\textwidth}{!}{ 
        \begin{tabular}{c|cccccc|cccc}
            \hline\hline
            & \multicolumn{6}{c|}{ID} & \multicolumn{4}{c}{OOD} \\
            & \multicolumn{2}{c}{Square} & \multicolumn{2}{c}{Triangle} & \multicolumn{2}{c|}{Hexagon} & \multicolumn{2}{c}{Pentagon} & \multicolumn{2}{c}{Round} \\ 
            & Suc         & Step         & Suc          & Step          & Suc          & Step          & Suc          & Step          & Suc         & Step        \\ \hline
            \rowcolor[HTML]{DAEFF9} VTLA & 95         & 4.31         & 95         & 3.94         & 95           & 3.52           & 100          & 1.85            & 100        & 5.2          \\ \hline\hline
        \end{tabular}
        }
    \end{minipage}%
    \vspace{-10pt}
\end{table*}

In the real-world insertion experiment, we comprehensively evaluate the Sim2Real capabilities of VTLA and other LLM-based methods (TLA and VLA). 
Each insertion task is conducted 20 times on the real robot.
First, we test the ability of VTLA model, trained entirely in simulation, to handle varying assembly clearances in real-world insertion tasks. The results in Tab.~\ref{tab_e3_1} show that as the assembly clearance decreases, 
the average assembly steps of VTLA increases from 1.6 to 4.3.
Despite this increased task difficulty, VTLA maintains a success rate above 95\%. We further examine the generalization ability of VTLA to different peg-hole shapes in real scenarios. The results in Tab. \ref{tab_e3_2} show that VTLA achieves a 100\% assembly success rate on OOD peg shapes, even slightly outperforming its performance on ID peg shapes.

\begin{wraptable}{r}{5.5cm}
    \setlength{\abovecaptionskip}{-0.5cm}
    \caption{Comparison of different methods on the peg-in-hole task\textbf{ with 0.6 mm clearance} in the real world.}
    \label{tab_e3_3}
    \centering
    \setlength{\tabcolsep}{0.7mm}{
    \begin{tabular}{ccccc}
    \hline\hline
    \multirow{2}{*}{Method}                & \multicolumn{2}{c}{Triangle} & \multicolumn{2}{c}{Pentagon} \\
                    & Suc          & Step          & Suc          & Step          \\ \hline
    VLA~\cite{qwen2-vl}             & 90           & 4.06          & \textbf{100}          & 2.3           \\
    TLA~\cite{hao2025tla}             & 30           & 2.00             & 40           & 1.88          \\
    \rowcolor[HTML]{DAEFF9} \textbf{VTLA (Ours)} & \textbf{95}           & 3.94          & \textbf{100}          & 1.85          \\ \hline\hline
    \end{tabular}
    }
\vspace{-10pt}
\end{wraptable}

\begin{figure*}[h] 
    \setlength{\abovecaptionskip}{0cm}
    \centering
    \includegraphics[width=\textwidth]{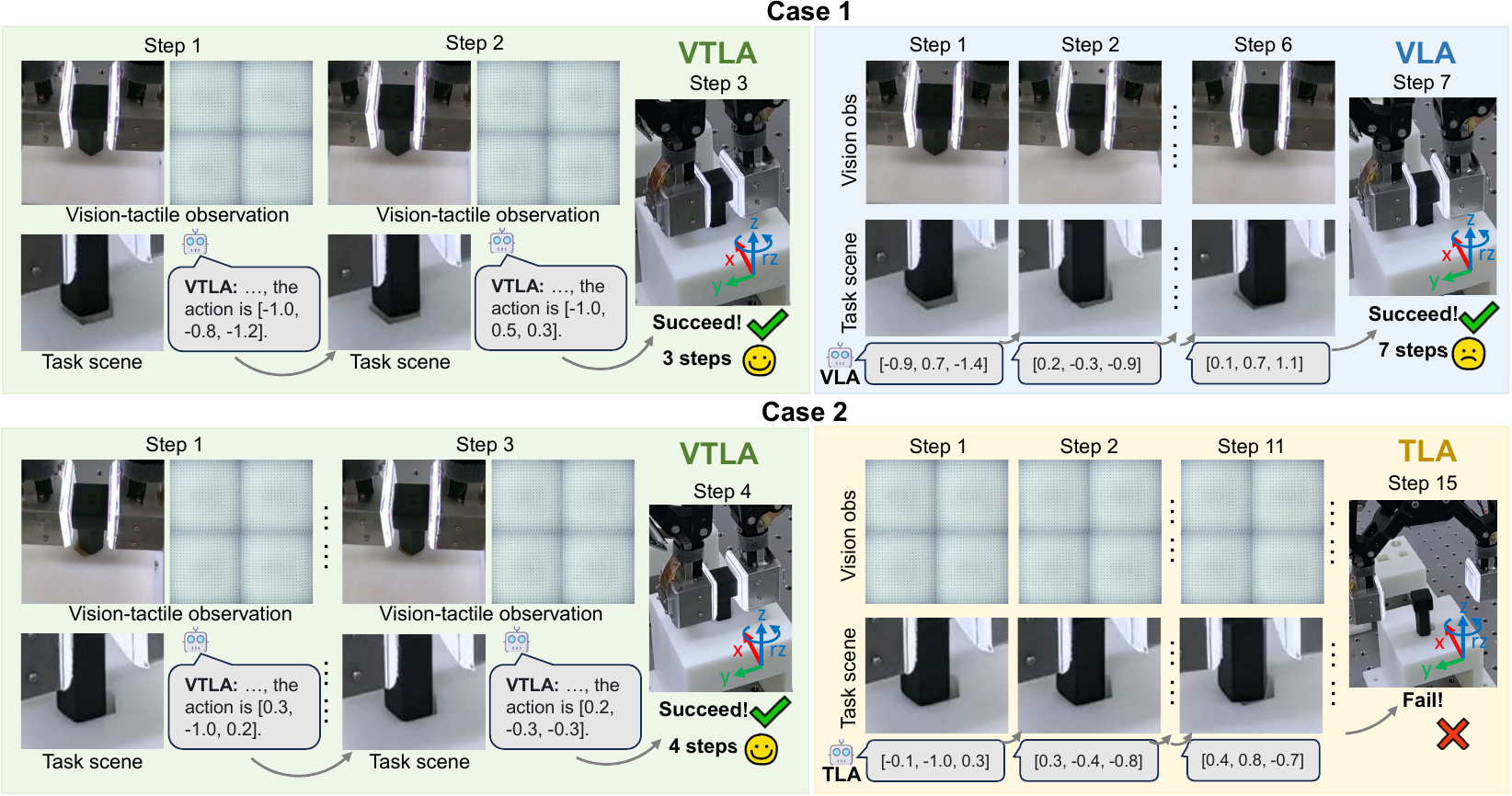}
    \caption{Snapshots of real-world insertion using the proposed VTLA model and baseline methods (VLA and TLA). The initial state of the task is consistent, and the assembly clearance is 0.6 mm.}
    \label{exp-real}
    \vspace{-5pt}
\end{figure*}

Compared with different LLM-based methods, the results in Tab. \ref{tab_e3_3} show that both VTLA and VLA model with vision modalities achieve good performance (over 90\%), but VTLA model with fusion of vision and tactile achieves better insertion efficiency (1.85 steps vs 2.3 steps in Tab. \ref{tab_e3_3}, Fig. \ref{exp-real}-Case 1).
Moreover, we find that the TLA model with only tactile observations faces a larger Sim2Real gap, with a success rate of only 30-40\%, which is only half of that in simulation. 
As shown in the Case 2 of Fig. \ref{exp-real}, under the same initial task states, VTLA completes the insertion within 3 steps, whereas TLA attempts 15 times but still fails to identify the correct direction, ultimately resulting in task failure. We suggest that exploring more advanced Sim2Real transfer methods may improve the real-world success rate of TLA model and potentially enable VTLA to handle more challenging insertion tasks.

\section{Conclusion}
\label{sec:conclusion}

	This study introduces the VTLA model, a vision-tactile-language-action model designed for contact-rich insertion manipulation tasks. The VTLA model is able to learn generalized visual-tactile skills based on language through a cross-modal fine-tuning process. 
    Experimental results show that in the challenging fingertip peg-in-hole task, VTLA significantly outperforms traditional imitation learning methods and surpasses both TLA and VLA models, achieving an assembly success rate exceeding 90\%. 
    The VTLA model 
    demonstrates strong task generalization capabilities across different assembly clearances and peg shapes. In addition, we also find that VTLA exhibits promising Sim2Real transfer performance, and the model trained using only simulated data can achieve an assembly success rate of 95\% in the real-world insertion task.

\clearpage
\textbf{Limitations}
Despite these encouraging results, VTLA has some areas for enhancement. One key aspect is aligning the tactile modality with the language modality, particularly regarding the semantic information of contact states for contact-rich manipulation tasks. In this paper, we utilize an off-the-shelf vision encoder to represent tactile inputs, which might led to a loss of the unique features inherent in the tactile modality. 
Future work on dedicated tactile-language alignment is necessary.
Another area for improvement is the fusion of visual and tactile modalities. While the deep integration of visual and tactile signals and features has been explored in proprietary models, corresponding research for LLMs remains limited. We believe this represents a promising topic for future work and encourage the community to explore these challenges together.
\bibliography{example}  

\clearpage
\appendix
\section*{Appendix}
\setcounter{page}{1}

This supplementary material provides additional details on the proposed method and experimental results that could not be included in the main manuscript due to page limitations.
Specifically, this appendix is organized as follows.
\begin{itemize}[left=1em]
    \item Sec.~\ref{secA} presents additional details of domain randomization on simulated dataset. 
    \item Sec.~\ref{secB} presents details of real-world robot setup for insertion task.
    \item Sec.~\ref{secC} presents a comparison between VTLA and VLA model under poor lighting condition.
\end{itemize}

\section{Domain Randomization on Simulated Dataset}
\label{secA}

To enhance zero-shot Sim2Real transfer performance of VTLA model, domain randomization techniques are employed during data generation to vary parameters in the simulation environment, task configurations, and both visual and tactile observations. Tab. \ref{tab:domain_random} shows the randomization parameters in details. 

\begin{table}[h]
	\centering
	\caption{The settings of domain randomization parameters.}
	\renewcommand\arraystretch{1.2}
	\label{tab:domain_random}
    \normalsize
	\begin{tabular}{c c c}
		\hline\hline
		& Parameter names                   & Distribution    \\ \hline
		\multirow{3}{*}{Physical}       & Young’s modulus (Pa)             & U(1.0e5, 5.0e5) \\
		& Poisson ratio                    & U(0.3, 0.48)    \\ 
		& Friction coefficient 				& U(0.2, 0.7) \\ \hline
		\multirow{3}{*}{Task-related}           & Peg offset x in gripper (mm)                & U(-1.0, 1.0)    \\
		& Peg offset z in gripper (mm)                & U(-1.0, 1.0)    \\
		& Contact depth (mm)               & U(0.6, 0.9)     \\ \hline
		\multirow{1}{*}{Tactile images} 
        & \thead{\normalsize{Color jittering (adjustment of brightness,}\\\normalsize{contrast, saturation, and hue)}}   &  / \\ \hline
        \multirow{7}{*}{Vision images} & Direction of environmental light source  & Any direction in 3D space     \\
		& Intensity of environmental light source & U(0.2, 0.6)      \\ 
        & Image scale transformation & U(0.9, 1.1) \\ 
        & Image translation transformation (pixels) & U(-10, 10) \\
        & Image rotation transformation (deg) & U(-3, 3) \\ 
        & Image shearing transformation (deg) & U(-3, 3) \\ 
        & Color jittering, Gaussian noise, motion blur, etc. & / \\  \hline\hline
	\end{tabular}
	\begin{tablenotes}
		\item[1] U(low, high) indicates a uniform distribution.
	\end{tablenotes}
\end{table}

\begin{itemize}
    \item \textbf{Physical parameters}: Young's modulus, Poisson ratio, and the friction coefficient are randomized. This is motivated by the fact that aging of the silicone layer in the visuotactile sensor can alter these parameters over time, and accurately determining their values in the real world is highly challenging. Furthermore, the friction coefficient in physical environments is influenced by numerous factors, such as material properties, ambient humidity, and so on, making precise modeling in simulation inherently difficult.
    \item \textbf{Task-related}: In real-world scenarios, the robot's grasping position and applied force on the peg may vary across episodes. To improve the policy's adaptability, we randomize the in-hand position of the peg and the gripper width during task initialization in the simulation environment.
    \item \textbf{Tactile images}: In order to reduce the sim-real gap of tactile images, color jittering is applied to the simulated tactile images, including adjustments of brightness, contrast, saturation, and hue.
    \item \textbf{Vision images}: Domain randomization is applied to both environmental lighting and vision images. At the start of each task, three light sources from different directions are randomly configured in the environment, with their intensities individually randomized. To reduce the sim-real gap caused by discrepancies between the wrist camera poses in the real and simulated environments, the vision images are randomly scaled, translated, rotated, and sheared. Additionally, color jittering, Gaussian noise, and motion blur are applied to further improve the generalization of VTLA model.
\end{itemize}

\section{Real-World Robot Setup for Insertion Task}
\label{secB}

\begin{figure*}[h] 
    \setlength{\abovecaptionskip}{0cm}
    \centering
    \includegraphics[width=\textwidth]{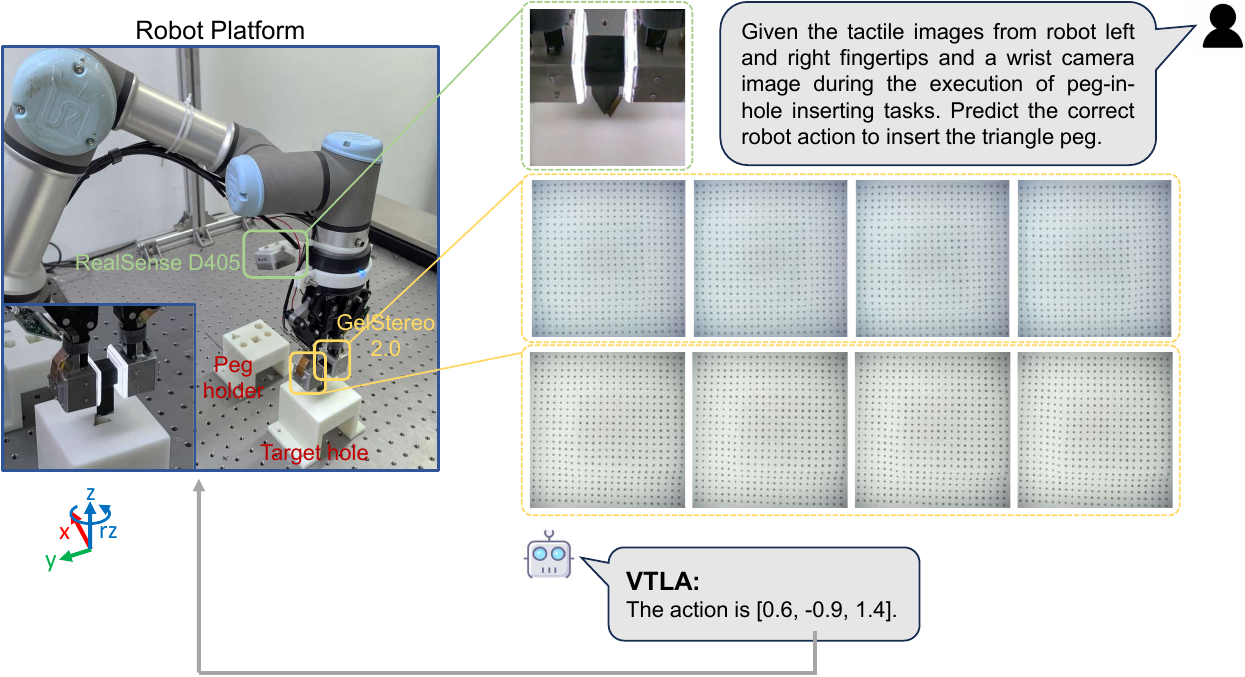}
    \caption{The insertion task setup in the real world. The left part shows the robot platform, and the right part shows the real visual-tactile observations and a dialogue round.}
    \label{exp-robot-setup}
    \vspace{-5pt}
\end{figure*}

We use a 6-DoF UR3 robot arm with a Robotiq 2F-85 gripper for real-world insertion experiments, as shown in Fig. \ref{exp-robot-setup}. An Intel RealSense D405 camera is mounted on the wrist to capture images, while two GelStereo 2.0 sensors \cite{zhang2023gelstereo} on the gripper's fingertips obtain tactile observations during peg-hole collisions. The robot begins by grasping the peg from a peg holder and approaching the target hole with a randomized pose. The misalignment between the peg and hole is sampled from a range of $-2.5$ to $2.5$ mm along the x-axis and y-axis, and from $-5^\circ$ to $5^\circ$ in rotation around the z-axis.
The gripper then moves down to attempt insertion.
The vision image is taken at the moment when the peg and hole contact. The tactile images sequences are recorded during peg-hole collisions, with each tactile sensor capturing a sequence of 4 frames.
The tactile sensing program is running on ROS at a rate of 20 PFS. Subsequently, the visual-tactile observations and a language instruction are fed into the VTLA model to generate a robot action. This process is repeated iteratively until the task is either successfully completed or deemed a failure.

\section{VTLA vs. VLA under Poor Lighting Condition}
\label{secC}

\begin{figure*}[t] 
    \setlength{\abovecaptionskip}{0cm}
    \centering
    \includegraphics[width=0.7\textwidth]{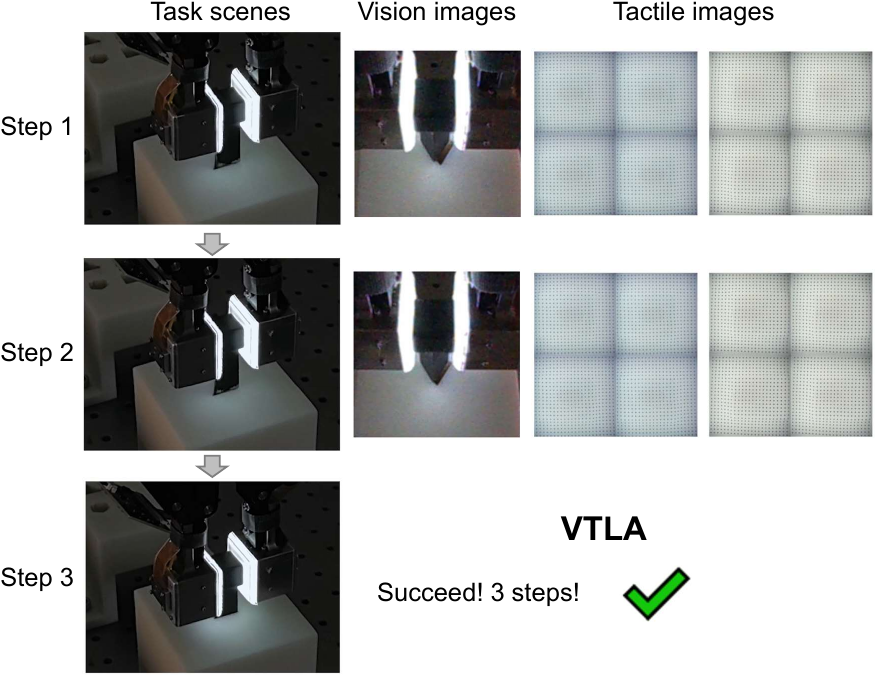}
    \caption{Snapshots of real-world insertion using the proposed \textbf{VTLA model} under poor lighting condition. The assembly clearance is 0.6 mm.}
    \label{dark_vtla}
    \vspace{-5pt}
\end{figure*}

\begin{figure*}[t] 
    \setlength{\abovecaptionskip}{0cm}
    \centering
    \includegraphics[width=0.88\textwidth]{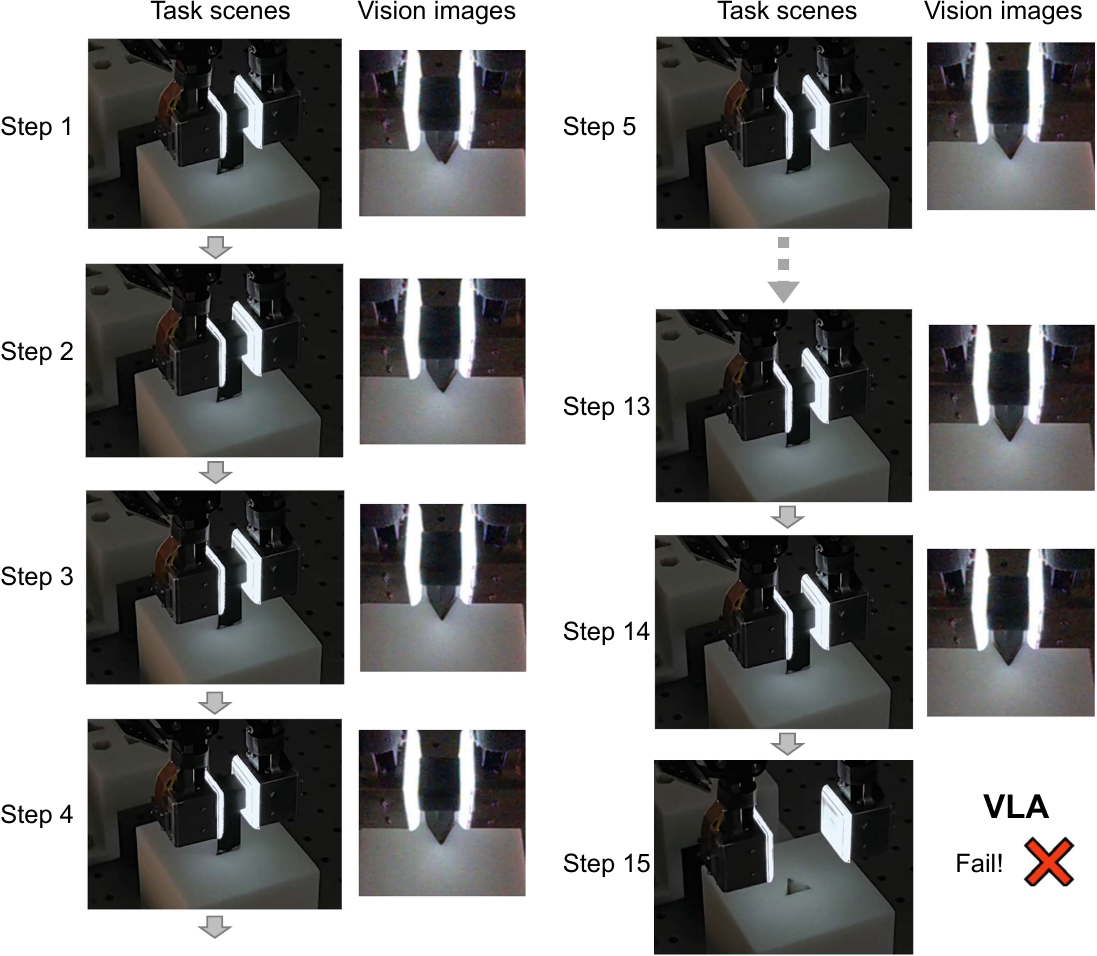}
    \caption{Snapshots of real-world insertion using the \textbf{VLA model} under poor lighting condition. The assembly clearance is 0.6 mm. The initial peg-hole misalignment is consistent with that in Fig. \ref{dark_vtla}.}
    \label{dark_vla}
    \vspace{-5pt}
\end{figure*}

We compare the performance of the VTLA and VLA models on the peg-in-hole assembly task under dim lighting conditions. The procedure of peg insertion is illustrated in Fig. \ref{dark_vtla} and Fig. \ref{dark_vla}, respectively. 
Under dim lighting conditions, the quality of the vision image is significantly degraded (comparing the vision images in Fig. \ref{exp-robot-setup} and Fig. \ref{dark_vtla}), making it difficult to recognize the hole position in the visual modality (steps 3–14 in Fig. \ref{dark_vla}). 
The VTLA model can successfully perform the insertion task under poor lighting condition, whereas VLA model struggles to complete the task.
This result demonstrates that the VTLA model is effective in leveraging both visual and tactile observations, and improves the success rate of contact-rich manipulation tasks.

\end{document}